\pdfoutput=1
\documentclass[11pt]{article}

\usepackage[final]{acl}

\usepackage{times}
\usepackage{latexsym}
\usepackage{multirow}
\usepackage{graphicx}
\usepackage{booktabs}
\usepackage{enumitem}
\usepackage{algorithm}
\usepackage{algorithmic}
\usepackage{amsmath}
\usepackage{bm}
\usepackage{colortbl}
\usepackage{longtable}
\usepackage{array}
\usepackage{hyperref}
\usepackage{tabularx}
\usepackage{pifont}
\usepackage{mydef}
\tikzset{forked edges/.style={}} % Define dummy 'forked edges' key to prevent the error
\usepackage{natbib}
\usepackage{siunitx}     % 数字对齐

\usepackage{multicol}
\usepackage{xcolor}
\setlist{nosep, leftmargin=12pt} % 紧凑列表格式
\usepackage[T1]{fontenc}
\usepackage[utf8]{inputenc}

\usepackage{microtype}

\usepackage{inconsolata}

\title{
A Survey on Efficient Large Language Model Training:\\ From \underline{Data-centric} Perspectives
}

\makeatletter
\renewcommand{\@fnsymbol}[1]{\textdagger}
\makeatother
\author{
Junyu Luo\textsuperscript{\rm $1$}, 
Bohan Wu\textsuperscript{\rm $1$}, 
Xiao Luo\textsuperscript{\rm $2$}\thanks{Corresponding authors.}, 
Zhiping Xiao\textsuperscript{\rm $3$}\footnotemark[1],
Yiqiao Jin\textsuperscript{\rm $4$},
Rong-Cheng Tu\textsuperscript{\rm $5$}, \\
\textbf{Nan Yin}\textsuperscript{\rm $6$}, 
\textbf{Yifan Wang}\textsuperscript{\rm $7$},
\textbf{Jingyang Yuan}\textsuperscript{\rm $1$}, 
\textbf{Wei Ju}\textsuperscript{\rm $1$}, 
\textbf{Ming Zhang}\textsuperscript{\rm $1$}\footnotemark[1] \\ 
{\textsuperscript{\rm $1$} State Key Laboratory for Multimedia Information Processing,} \\
{School of Computer Science, PKU-Anker LLM Lab, Peking University} \\
{\textsuperscript{\rm $2$} University of California, Los Angeles} 
{\textsuperscript{\rm $3$} University of Washington}\\
{\textsuperscript{\rm $4$} Georgia Institute of Technology} 
{\textsuperscript{\rm $5$} Nanyang Technological University}\\
{\textsuperscript{\rm $6$} HKUST} 
{\textsuperscript{\rm $7$} University of International Business and Economics} 
\\
{\tt \url{https://github.com/luo-junyu/Awesome-Data-Efficient-LLM}}\\
}

\definecolor{hidden-draw}{RGB}{0,0,0}
\definecolor{hidden-pink}{RGB}{255,245,247}
\definecolor{hidden-blue}{RGB}{74,144,226}
\definecolor{hidden-orange}{RGB}{234,148,36}
\definecolor{hidden-green}{RGB}{39,167,0}

\definecolor{LightCyan}{rgb}{0.88,1,1}

\newcommand{\paratitle}[1]{\noindent\emph{\textbf{#1}}}

\begin{document}
\maketitle
\begin{abstract}

Post-training of Large Language Models (LLMs) is crucial for unlocking their task generalization potential and domain-specific capabilities. However, the current LLM post-training paradigm faces significant data challenges, including the high costs of manual annotation and diminishing marginal returns on data scales. Therefore, achieving data-efficient post-training has become a key research question. In this paper, we present the first systematic survey of data-efficient LLM post-training from a data-centric perspective. We propose a taxonomy of data-efficient LLM post-training methods, covering data selection, data quality enhancement, synthetic data generation, data distillation and compression, and self-evolving data ecosystems. We summarize representative approaches in each category and outline future research directions. By examining the challenges in data-efficient LLM post-training, we highlight open problems and propose potential research avenues. We hope our work inspires further exploration into maximizing the potential of data utilization in large-scale model training.

\end{abstract}

\section{Introduction}

Large Language Models~(LLMs) post-training has emerged as a crucial stage for unlocking their domain adaptation capabilities and task generalization potential~\cite{luo2025large}. This phase has effectively enhanced models' abilities in long-context reasoning~\cite{zelikman2022star,yuan2024hybrid}, human alignment~\cite{rafailov2024direct}, instruction tuning~\cite{zhang2023instruction}, and domain-specific adaptation~\cite{cheng2024adapting}.

\begin{figure}[t]
    \centering
    \includegraphics[width=\linewidth]{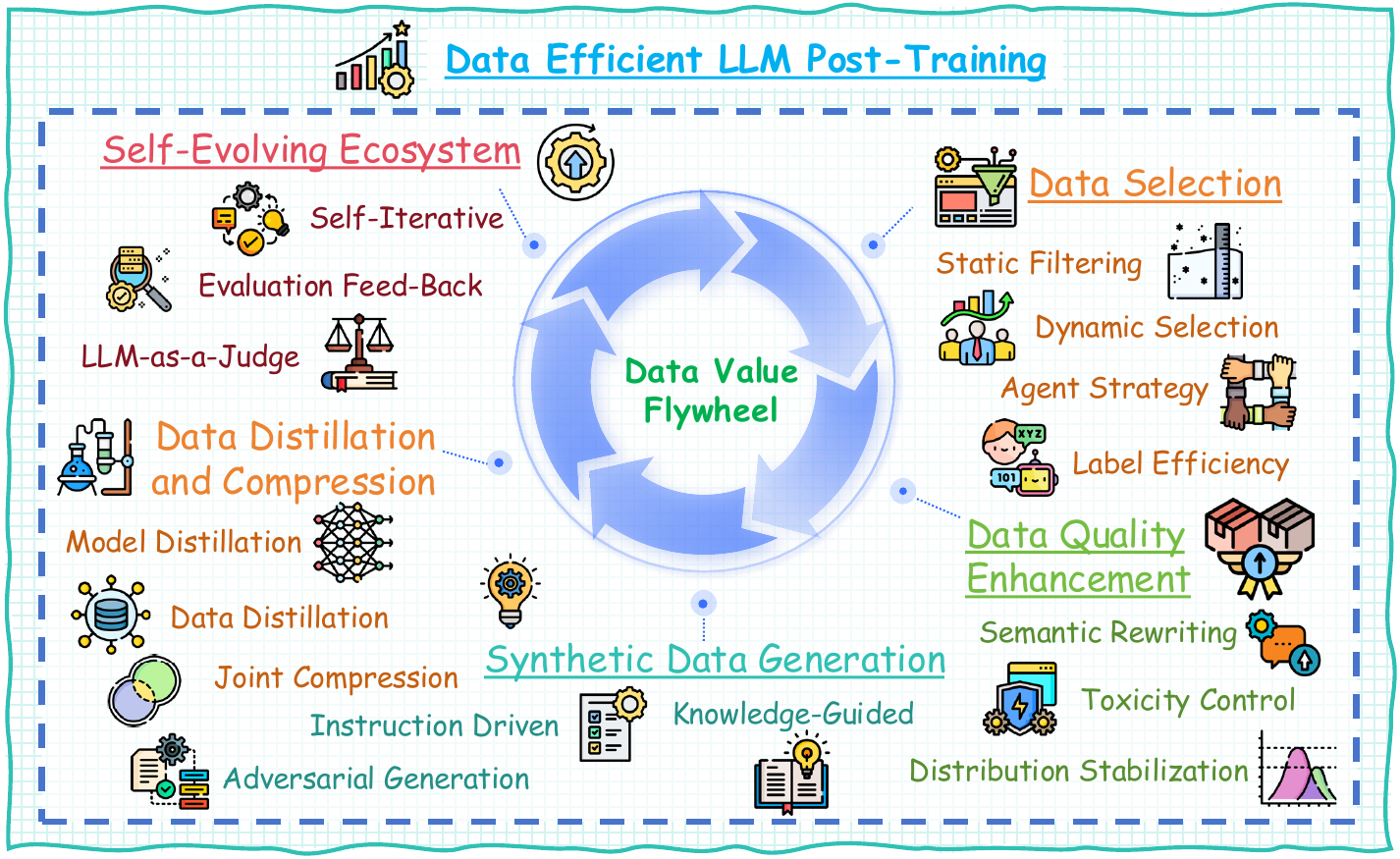}
    \caption{Illustration of the data flywheel in \textbf{Data-Efficient LLM Post-Training}, depicting the iterative cycle of data selection, data quality enhancement, synthetic data generation, knowledge distillation, and self-evolving data ecosystems to maximize model performance with minimal data requirements. }
    \vspace{-4mm}
    \label{fig:data-flywheel}
\end{figure}

During the LLM post-training phase, data is the essential driver of model evolution. However, the current paradigm faces a severe \textit{data dilemma}: the cost of manually annotating high-quality data is rapidly growing, while simply scaling data volume yields diminishing returns. Moreover, static datasets inherently limit models from adapting to evolving real-world knowledge. The linear dependency between data volume and model performance fundamentally stems from the inefficient data usage in traditional post-training paradigms. The recent success of DeepSeek-R1~\cite{guo2025deepseek}, which leverages reinforcement learning for data-efficient post-training, further demonstrates the effectiveness and necessity of data-efficient approaches in achieving superior LLM performance. Our work establishes the \textbf{first systematic survey on data-efficient training} of LLMs, providing a unified, taxonomized framework to address the fragmented research landscape. Our survey reveals that breaking through efficiency bottlenecks requires establishing value extraction across the lifecycle, rather than merely expanding data scale.

Researchers have explored various approaches to fully exploit the data potential in LLM post-training~\cite{jeong2024llm,wang2024data,pan2024preparing}. While these methods have made notable progress in improving data efficiency, the field still lacks a comprehensive review. In this paper, we provide a comprehensive survey of data-efficient LLM post-training from a data-centric perspective. Specifically, we introduce the concept of a \textit{data value flywheel}~(as illustrated in Figure~\ref{fig:data-flywheel}), which consists of five key components: data selection, data quality enhancement, synthetic data generation, data distillation and compression, and self-evolving data ecosystems. Using this framework, we present a taxonomy of existing work, summarize key components, and identify promising research directions.  We hope our work serves as both a useful roadmap for newcomers and a guide for future advancements in the field.

\paratitle{Differences from previous surveys.} While several surveys have explored a few aspects of LLMs post-training, including data selection~\cite{wang2024survey}, synthetic data generation~\cite{long2024llms,tan2024large}, model self-feedback~\cite{liang2024internal,pan2023automatically}, self-evolution~\cite{tao2024survey}, trustworthiness~\cite{liu2023trustworthy}, and time-efficiency~\cite{wan2023efficient}, these studies primarily focus on individual aspects rather than a holistic perspective. Our survey fills this gap by systematically examining these methods through the lens of data efficiency, offering critical insights into maximizing data value extraction.

\definecolor{softblue}{RGB}{220,230,242}    % 柔和的蓝色
\definecolor{softgreen}{RGB}{226,239,218}   % 柔和的绿色
\definecolor{softgray}{RGB}{229,224,236}  % 柔和的紫色
\definecolor{softgray}{RGB}{255,242,204}  % 柔和的黄色
\definecolor{softred}{RGB}{242,220,219}     % 柔和的红色
\definecolor{softgray}{RGB}{240,240,240}     % 柔和的灰色

\tikzstyle{leaf}=[draw=black, %边框
    rounded corners,minimum height=1em,
    text width=22.50em, edge=black!10, 
    %fill=hidden-orange!40,
    text opacity=1, 
    align=left,
    fill opacity=.3,  text=black,font=\scriptsize,
    inner xsep=5pt, inner ysep=3pt,
    ]
\tikzstyle{leaf1}=[draw=black, %边框
    rounded corners,minimum height=1em,
    text width=6.5em, edge=black!10, 
    text opacity=1, align=center,
    fill opacity=.5,  text=black,font=\scriptsize,
    inner xsep=3pt, inner ysep=3pt,
    ]
\tikzstyle{leaf2}=[draw=black, %边框
    rounded corners,minimum height=1em,
    text width=6.5em, edge=black!10, 
    text opacity=1, align=center,
    fill opacity=.8,  text=black,font=\scriptsize,
    inner xsep=3pt, inner ysep=3pt,
    ]
\tikzstyle{leaf3}=[draw=black, %边框
    rounded corners,minimum height=1em,
    text width=6em, edge=black!10, 
    text opacity=1, align=center,
    fill opacity=.8,  text=black,font=\scriptsize,
    inner xsep=3pt, inner ysep=3pt,
]
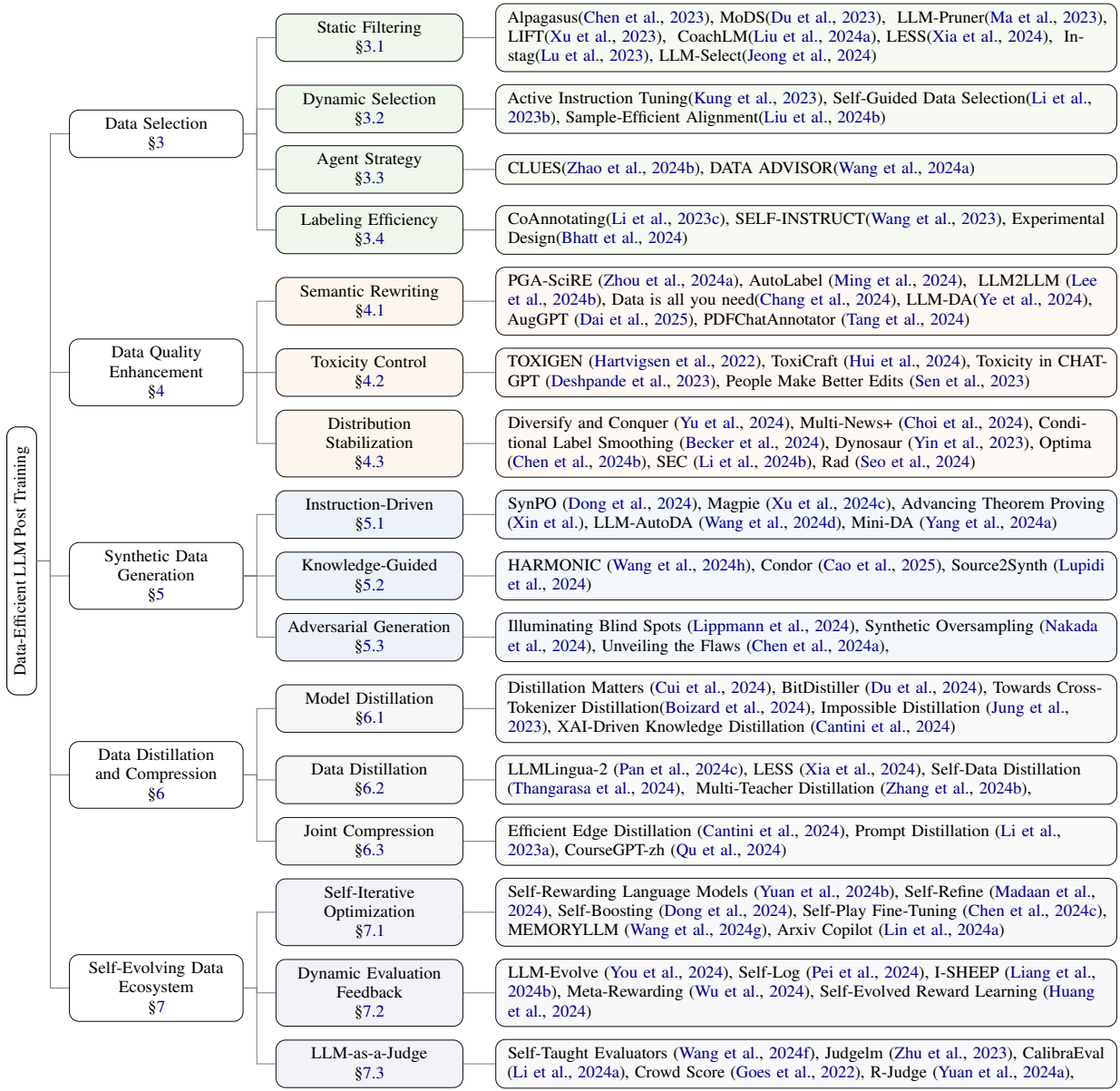
\begin{figure*}[t]
\centering
\begin{forest}
  for tree={
  forked edges,
  grow=east,
  reversed=true,
  anchor=center,
  parent anchor=east,
  child anchor=west,
  base=center,
  font=\scriptsize,
  rectangle,
  draw=black, %hiddendraw 所有边框
  edge=black!50, 
  rounded corners,
  minimum width=2em,
  s sep=5pt,
  inner xsep=3pt,
  inner ysep=1pt
  },
  where level=1{text width=4.5em}{},
  where level=2{text width=6em,font=\scriptsize}{},
  where level=3{font=\scriptsize}{},
  where level=4{font=\scriptsize}{},
  where level=5{font=\scriptsize}{},
  [Data-Efficient LLM Post Training,rotate=90,anchor=north,inner xsep=8pt,inner ysep=3pt,edge=black!50,draw=black
    [Data Selection \\ \S \ref{sec:data_selection}, edge=black!50, leaf3,
      [Static Filtering \\ \S \ref{sec:static_filtering}, leaf1, fill=softgreen,
          [Alpagasus\cite{chen2023alpagasus}{, }MoDS\cite{du2023mods}{, }          LLM-Pruner\cite{ma2023llm}{, }LIFT\cite{xu2023rethinking}{, }          CoachLM\cite{liu2024coachlm}{, }LESS\cite{xia2024less}{, }
          Instag\cite{lu2023instag}{, }LLM-Select\cite{jeong2024llm},
            ,leaf,fill=softgreen]
      ]
      [Dynamic Selection \\ \S \ref{sec:dynamic_selection}, leaf1, fill=softgreen,
        [Active Instruction Tuning\cite{kung2023active}{, }Self-Guided Data Selection\cite{li2023quantity}{, }Sample-Efficient Alignment\cite{liu2024sample},leaf,fill=softgreen]
      ]
      [Agent Strategy \\ \S \ref{sec:agent_strategy}, leaf1, fill=softgreen,
        [CLUES\cite{zhao2024clues}{, }DATA ADVISOR\cite{wang2024data},leaf,fill=softgreen]
      ]
      [Labeling Efficiency \\ \S \ref{sec:labeling_efficiency}, leaf1, fill=softgreen,
        [CoAnnotating\cite{li2023coannotating}{, }SELF-INSTRUCT\cite{wang2023self}{, }Experimental Design\cite{bhatt2024experimental},leaf,fill=softgreen]
      ]
    ]
    [Data Quality\\ Enhancement \\ \S \ref{sec:data_quality}, edge=black!50, leaf3, 
      [Semantic Rewriting \\ \S \ref{sec:semantic_rewriting}, leaf1, fill=myyellow,
        [PGA-SciRE \cite{zhou2024pga}{, }AutoLabel \cite{ming2024autolabel}{, } LLM2LLM \cite{lee2024llm2llm}{, }Data is all you need\cite{chang2024data}{,} LLM-DA\cite{ye2024llm}{, }AugGPT \cite{dai2025auggpt}{, }PDFChatAnnotator \cite{tang2024pdfchatannotator},leaf,fill=myyellow]
      ]
      [Toxicity Control \\ \S \ref{sec:toxicity_control}, leaf1, fill=myyellow,
        [TOXIGEN \cite{hartvigsen2022toxigen}{, }ToxiCraft \cite{hui2024toxicraft}{, }Toxicity in CHATGPT \cite{deshpande2023toxicity}{, }People Make Better Edits \cite{sen2023people} ,leaf,fill=myyellow]
      ]
      [Distribution Stabilization \\ \S \ref{sec:distribution_stabilization}, leaf1, fill=myyellow,
        [Diversify and Conquer \cite{yu2024diversify}{, }Multi-News+ \cite{choi2024multi}{, }Conditional Label Smoothing \cite{becker2024conditional}{, }Dynosaur \cite{yin2023dynosaur}{, }Optima \cite{chen2024optima}{, }SEC \cite{li2024synergized}{, }Rad \cite{seo2024retrieval},leaf,fill=myyellow]
      ]
    ]
    [Synthetic Data\\ Generation \\ \S \ref{sec:synthetic_data_generation}, edge=black!50, leaf3, 
      [Instruction-Driven \\ \S \ref{sec:instruction_driven_synthetic_data_generation}, leaf1, fill=softblue,
        [SynPO \cite{dong2024self}{, }Magpie \cite{xu2024magpie}{, }Advancing Theorem Proving \cite{xin24advancing}{, }LLM-AutoDA \cite{wangllm}{, }Mini-DA \cite{yang2024mini} ,leaf,fill=softblue]
      ]
      [Knowledge-Guided \\ \S \ref{sec:knowledge_guided_synthetic_data_generation}, leaf1, fill=softblue,
        [HARMONIC \cite{wang2024harmonic}{, }Condor \cite{cao2025condor}{, }Source2Synth \cite{lupidi2024source2synth} ,leaf,fill=softblue]
      ]
      [Adversarial Generation \\ \S \ref{sec:adversarial_generation}, leaf1, fill=softblue,[Illuminating Blind Spots \cite{lippmann2024illuminating}{, }Synthetic Oversampling \cite{nakada2024synthetic}{, }Unveiling the Flaws \cite{chen2024unveiling}{, },leaf,fill=softblue]]
    ]
    [Data Distillation \\ and Compression \\ \S \ref{sec:data_distillation_compression}, edge=black!50, leaf3, 
      [Model Distillation \\ \S \ref{sec:model_distillation}, leaf1, fill=softgray,
        [Distillation Matters \cite{cui2024distillation}{, }BitDistiller \cite{du2024bitdistiller}{, }Towards Cross-Tokenizer Distillation\cite{boizard2024towards}{, }Impossible Distillation \cite{jung2023impossible}{, }XAI-Driven Knowledge Distillation \cite{cantini2024xai} ,leaf,fill=softgray]
      ]
      [Data Distillation \\ \S \ref{sec:data_distillation}, leaf1, fill=softgray,
        [LLMLingua-2 \cite{pan2024llmlingua}{, }LESS \cite{xia2024less}{, }Self-Data Distillation \cite{thangarasa2024self}{, } Multi-Teacher Distillation \cite{zhang2024llm}{, } ,leaf,fill=softgray]
      ]
      [Joint Compression \\ \S \ref{sec:joint_compression}, leaf1, fill=softgray,
        [Efficient Edge Distillation \cite{cantini2024xai}{, }Prompt Distillation \cite{li2023prompt}{, }CourseGPT-zh \cite{qu2024coursegpt} ,leaf,fill=softgray]
      ]
    ]
    [Self-Evolving Data \\ Ecosystem \\ \S \ref{sec:self_evolving}, edge=black!50, leaf3, 
      [Self-Iterative \\ Optimization \\ \S \ref{sec:self_iterative_optimization}, leaf1, fill=mypurple,
        [Self-Rewarding Language Models \cite{yuan2024self}{, }Self-Refine \cite{madaan2024self}{, }Self-Boosting \cite{dong2024self}{, }Self-Play Fine-Tuning \cite{chen2024self}{, }MEMORYLLM \cite{wang2024memoryllm}{, }Arxiv Copilot \cite{lin2024arxiv} ,leaf,fill=mypurple]
      ]
      [Dynamic Evaluation \\ Feedback \\ \S \ref{sec:dynamic_evaluation_feedback}, leaf1, fill=mypurple,
        [LLM-Evolve \cite{you2024llm}{, }Self-Log \cite{pei2024self}{, }I-SHEEP \cite{liang2024sheep}{, }Meta-Rewarding \cite{wu2024meta}{, }Self-Evolved Reward Learning \cite{huang2024self} ,leaf,fill=mypurple]
      ]
      [LLM-as-a-Judge \\ \S \ref{sec:llm_as_a_judge}, leaf1, fill=mypurple,
        [Self-Taught Evaluators \cite{wang2024self}{, }Judgelm \cite{zhu2023judgelm}{, }CalibraEval \cite{li2024calibraeval}{, }Crowd Score \cite{goes2022crowd}{, }R-Judge \cite{yuan2024r}{, } ,leaf,fill=mypurple]
      ]
    ]
  ]
\end{forest}
\caption{A taxonomy of Data-Efficient LLM Post Training.}
\vspace{-4mm}
\label{fig:taxonomy_data_efficient_llm_post_training}
\end{figure*}

\section{Taxonomy}

This section categorizes data-efficient post-training methods for LLMs into five core methodologies:

\begin{itemize}[leftmargin=*]
    \item \textbf{Data Selection}: \textit{Filtering high-value subsets from raw data.} \ding{182} Static Filtering: Offline selection based on data properties; \ding{183} Dynamic Selection: Adjusting weights based on model uncertainty; \ding{184} Agent Strategy: Multi-model voting for reliable selection; \ding{185} Labeling Efficiency: Combining active learning and semi-supervised strategies for cost-effective sample coverage.

    \item \textbf{Data Quality Enhancement}: \textit{Improving the utility of existing data.} \ding{182} Semantic Rewriting: Enhancing expression diversity through semantic-preserving transformations and generating variants while maintaining original meaning; \ding{183} Toxicity Control: Correcting harmful content; \ding{184} Distribution Stabilization: Adjusting data characteristics for robustness

    \item \textbf{Synthetic Data Generation}: \textit{Creating new training data.} \ding{182} Instruction-Driven: Model-generated instruction-response pairs; \ding{183} Knowledge-Guided: Generation with structured knowledge; \ding{184} Adversarial Generation: Producing challenging samples.

    \item \textbf{Data Distillation and Compression}: \textit{Extracting core knowledge for efficient training.} \ding{182} Model Distillation: Transferring large model output distributions to smaller models while preserving key knowledge; \ding{183} Data Distillation: Extracting high information density samples to construct compact datasets equivalent to full-scale data; \ding{184} Joint Compression: Combining model architecture compression with data selection strategies for end-to-end efficiency optimization

    \item \textbf{Self-Evolving Data Ecosystem}: \textit{Building self-evolution mechanisms.} \ding{182} Self-Iterative Optimization: Using current model to generate data; \ding{183} Dynamic Evaluation Feedback: Real-time monitoring and adjustment; \ding{184} LLM-as-a-Judge: Feedback-Driven Data Optimization;
\end{itemize}

\begin{table}[t]
\centering
\setlength{\tabcolsep}{2pt}
\resizebox{\linewidth}{!}{
\begin{tabular}{lcccc}
\toprule
\textbf{Category} & \textbf{Data} & \textbf{Compute} & \textbf{Model} & \textbf{Data Value} \\
& \textbf{Dependency} & \textbf{Cost} & \textbf{Dependency} & \textbf{Mining} \\
\midrule
Data Selection & ++ & + & + & +++ \\
Quality Enhance. & ++ & ++ & ++ & ++ \\
Synthetic Generation & + & +++ & +++ & + \\
Distill. \& Compress. & + & + & +++ & +++ \\
Self-Evolving & + & +++ & +++ & +++ \\
\bottomrule
\end{tabular}
}
\caption{Comparison of different data-efficient post-training methods across key dimensions.}
\vspace{-4mm}
\label{tab:method_comparison}
\end{table}

Table~\ref{tab:method_comparison} compares the five methodologies across key dimensions, where more '+' indicates higher requirements or better performance. Data selection shows high data efficiency but requires quality source data. Quality enhancement maintains balanced requirements across dimensions. Synthetic generation and self-evolving approaches demand more compute and model resources but reduce dependency on original manually annotated datasets, as they primarily rely on teacher model outputs or self-generated data. Distillation methods excel in data efficiency while depending on model capabilities.

These five dimensions complement each other: selection filters quality data, enhancement improves utility, generation expands coverage, distillation concentrates knowledge, and self-evolution enables continuous improvement. Together, they pursue the goal of \textit{less data, higher returns}.

% achieve efficient post-training with minimal data requirements.
% These five dimensions form an organic whole: selection methods ensure data quality, enhancement methods improve data utility, generation methods expand data boundaries, distillation methods achieve knowledge concentration, and self-evolution systems establish continuous optimization mechanisms. Together, they drive the post-training process toward the goal of \textit{less data, higher returns}.

% \input{4_pre}
\section{Data Selection}\label{sec:data_selection}

\begin{figure}[t]
    \centering
    \includegraphics[width=1\linewidth]{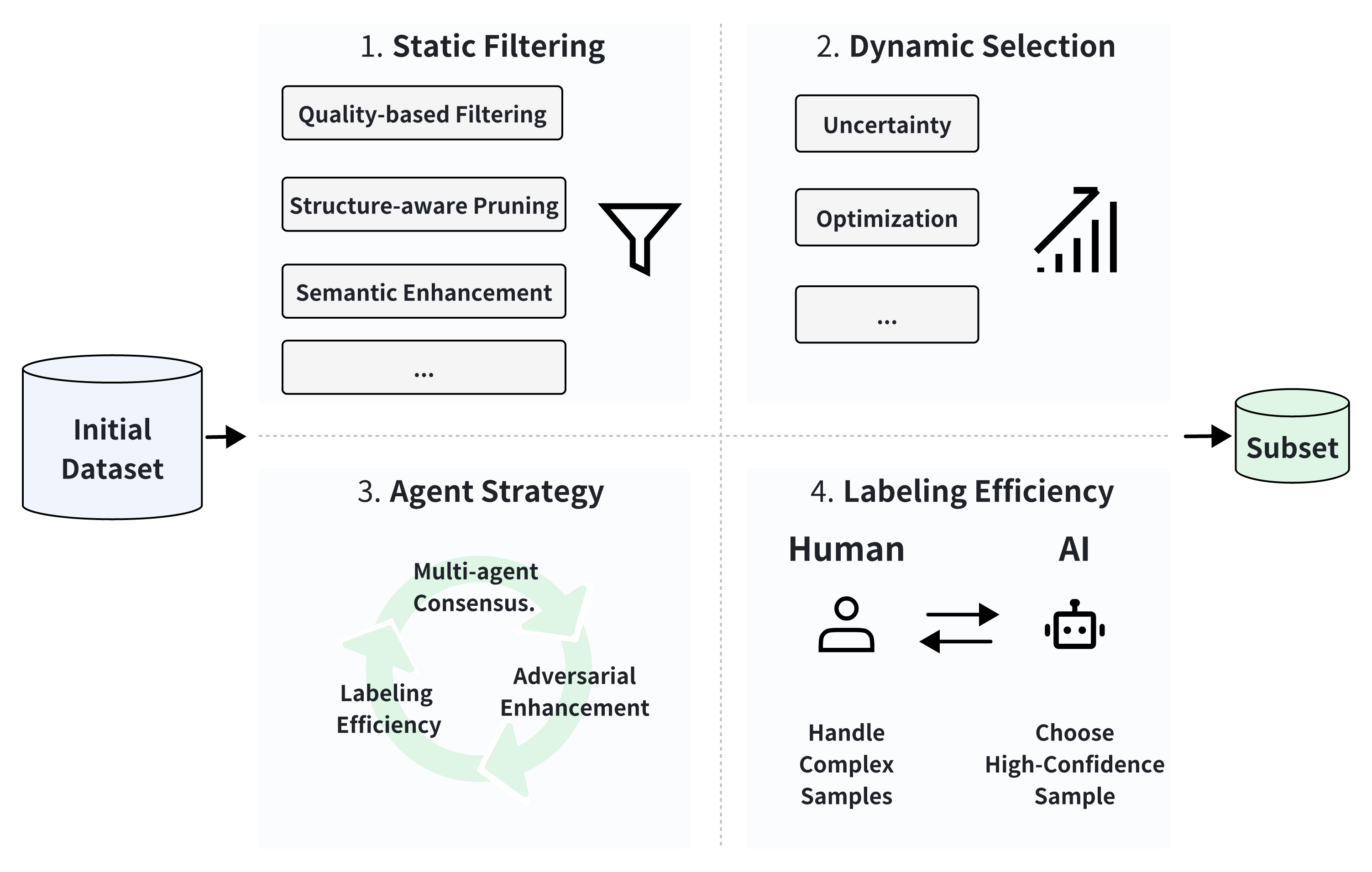}
    \vspace{-5mm}
    \caption{Overview of four major data selection approach categories: static filtering, dynamic selection, agent strategy, and labeling efficiency.}
    \vspace{-5mm}
    \label{fig:data-selection}
\end{figure}

Data selection is crucial for enhancing LLM post-training efficiency by identifying high-value data subsets. As shown in Figure~\ref{fig:data-selection}, we divide existing approaches into four dimensions: (1) static filtering based on inherent data properties, (2) dynamic selection that adapts during training, (3) agent strategy using collaborative mechanisms, and (4) labeling efficiency through human-AI collaboration.

\subsection{Static Filtering}\label{sec:static_filtering}
Static filtering evaluates inherent data properties offline to identify samples with high information density and representativeness.

\paratitle{Quality-based Filtering.} Alpagasus~\cite{chen2023alpagasus} achieves comparable performance using only 17\% of the original data through complexity-based filtering (instruction length, diversity, and perplexity). MoDS~\cite{du2023mods} employs multi-dimensional indicators and density peak clustering, while~\cite{kang2024get} uses KL-divergence-driven selection to align domain distributions. Information-theoretic approaches~\cite{kim2024measuring} leverage entropy metrics using negative log-likelihood and inverse word frequency to identify redundant samples. ActivePrune~\cite{azeemi2024language} implements two-stage pruning through n-gram perplexity scoring followed by quantized LLM evaluation. CoT-Influx~\cite{huang2023fewer} employs coarse-to-fine pruning for reasoning enhancement in mathematical tasks. LLM-Select~\cite{jeong2024llm} demonstrates that large language models can perform feature selection using only feature names and task descriptions, rivaling traditional data science tools.

% \paratitle{Semantic Enhancement.} LIFT~\cite{xu2023rethinking} enhance instruction quality through automatic revision. In recommendation systems, works like~\cite{lin2024data,lu2023instag,xia2024less,jeong2024llm} extend filtering methods using task-oriented scoring mechanisms and achieve better performance.

\paratitle{Semantic Enhancement.} LIFT~\cite{xu2023rethinking} enhances instruction quality through automatic revision. InsTag~\cite{lu2023instag} proposes fine-grained instruction tagging to analyze diversity and complexity in supervised fine-tuning datasets, demonstrating that model capability grows with more diverse and complex data.

% xia2024less
% lu2023instag
% jeong2024llm

\subsection{Dynamic Selection}\label{sec:dynamic_selection}
Dynamic methods adapt data weights by evaluating sample importance based on model feedback.

\paratitle{Uncertainty-driven Selection.} Active Instruction Tuning~\cite{kung2023active} prioritizes high-uncertainty tasks through prediction entropy. Self-Guided Data Selection uses Instruction Following Difficulty (IFD) to measure loss variance and eliminate easily learned examples~\cite{li2023quantity}.

\paratitle{Optimization-based Selection.} Sample-efficient alignment~\cite{liu2024sample} uses Thompson sampling to maximize contribution in preference alignment tasks. Compute-constrained data selection~\cite{yin2024compute} optimizes between data utility and computational cost. P3~\cite{yang2024p3} integrates policy-driven difficulty assessment with pace-adaptive selection and diversity promotion through Determinantal Point Process. LESS~\cite{xia2024less} employs optimizer-aware gradient similarity search with low-rank gradient features for targeted instruction tuning. In domain-specific applications, data pruning methods~\cite{lin2024data} use influence and effort scores to identify representative samples.

\subsection{Agent Strategy}\label{sec:agent_strategy}
Agent-based approaches leverage collaborative mechanisms for reliable selection.

\paratitle{Multi-agent Consensus.} Multi-agent methods like CLUES~\cite{zhao2024clues} implement multi-model voting mechanisms based on training dynamics and gradient similarity metrics.

\paratitle{Adversarial Enhancement.} Recent works like DATA ADVISOR~\cite{wang2024data} uses red-team agents for safety filtering, while Automated Data Curation~\cite{chen2024automated} optimizes data through generator-discriminator frameworks.

\subsection{Labeling Efficiency}\label{sec:labeling_efficiency}
These methods efficiently optimize annotation processes through iterative human-AI collaboration.

\paratitle{Human-AI Collaboration.} Methods like LLMaAA~\cite{zhang2023llmaaa} employ LLMs as annotators with uncertainty sampling. CoAnnotating~\cite{li2023coannotating} implements uncertainty-guided labor division between humans and AI.

\paratitle{Automated Generation.} SELF-INSTRUCT~\cite{wang2023self} enables autonomous self-generated instruction data, while~\cite{li2023one} uses one-shot learning for rapid sample identification.

\paratitle{Workflow Optimization.} Recent works establish scalabel efficient annotation workflows through adaptive experimental design~\cite{bhatt2024experimental} and systematic curation systems~\cite{pang2024improving}.

\subsection{Discussion}
Current data selection approaches face challenges in aligning static metrics with dynamic model requirements, managing computational complexity in optimization, and achieving cross-domain generalization. Future research points toward meta-learning-based selection frameworks, causal inference for sample analysis, and efficiency-aware optimization with hardware constraints, advancing data selection toward theoretical grounding.
\section{Data Quality Enhancement}\label{sec:data_quality}
\begin{figure}[t]
    \centering
    \includegraphics[width=\linewidth]{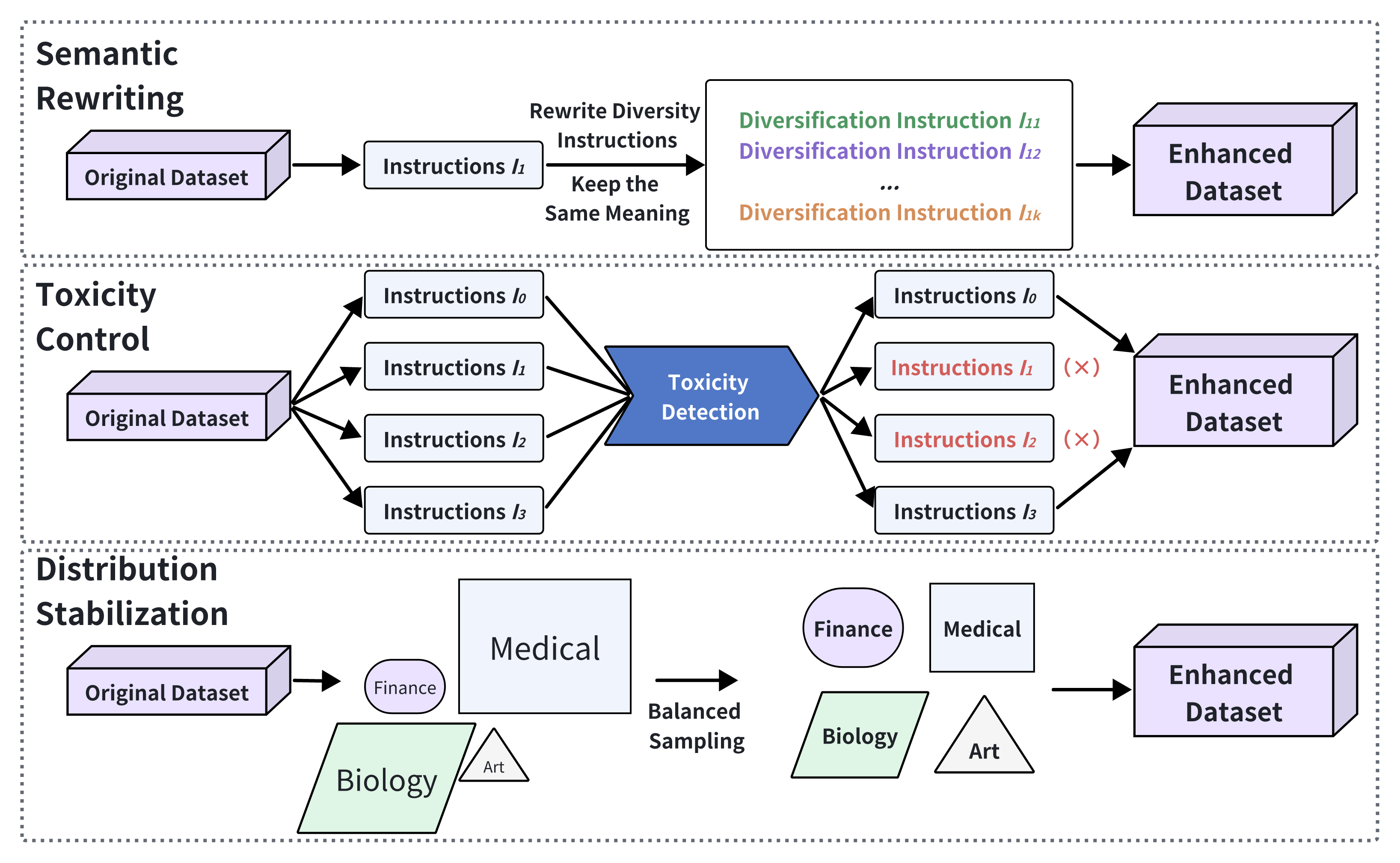}
    \vspace{-5mm}
    \caption{Three key approaches for data quality enhancement in LLM post-training: semantic rewriting for diversity, toxicity control for safety, and distribution stabilization for balanced representation.}
    \vspace{-5mm}
    \label{fig:data-quality-enhance}
\end{figure}

As illustrated in Figure~\ref{fig:data-quality-enhance}, enhancing data quality is critical for maximizing the effectiveness of LLM post-training~\cite{zhou2024survey}. Through semantic refinement, toxicity control, and distribution stabilization, researchers aim to improve the informativeness, safety, and robustness of training data. We categorize existing methods into three directions.

\subsection{Semantic Rewriting}\label{sec:semantic_rewriting}
Semantic rewriting focuses on augmenting data diversity while preserving original meaning through controlled transformations. This can be achieved through several key approaches:

\paratitle{Instruction Refinement.} CoachLM~\cite{liu2024coachlm} automatically revises complex instructions to reduce ambiguity, while~\cite{li2024empowering} uses structured prompt chains for paraphrase generation, enhancing model generalization across tasks.

\paratitle{Domain-Specific Augmentation.} Methods like~\cite{jia2024curriculum} use curriculum learning for metaphor detection, while PGA-SciRE~\cite{zhou2024pga} injects structured knowledge for scientific relation extraction, adapting models to specialized tasks.

\paratitle{Automated Enhancement.} AutoLabel~\cite{ming2024autolabel} seamlessly integrates human feedback for quality rewriting, while LLM2LLM~\cite{lee2024llm2llm} iteratively improves low-confidence samples. LANCE~\cite{wang2024language} enables LLMs to autonomously generate, clean, review, and annotate data, serving as continuous self-evolving data engineers. Recent studies extensively explore human-AI collaboration~\cite{chung2023increasing} and various data types: text~\cite{dai2025auggpt}, tabular~\cite{banday2024role}, and multimodal~\cite{tang2024pdfchatannotator}. LLM-DA~\cite{ye2024llm} employs contextual rewriting strategies with entity-level replacements for few-shot NER tasks, while~\cite{zhang2025data} leverages lightweight LLM generation and tree hybridization for cross-domain parsing augmentation.

\subsection{Toxicity Control}\label{sec:toxicity_control}
Mitigating harmful content is crucial for data quality enhancement. Recent approaches focus on detection, benchmarking, and human collaboration:

\paratitle{Detection Frameworks.} Methods like~\cite{zhang2024efficient} effectively distill toxicity knowledge into compact detectors, while~\cite{wang2022toxicity} strategically leverages generative prompts for zero-shot toxicity classification across diverse tasks.

\paratitle{Adversarial Benchmarking.} Frameworks such as TOXIGEN~\cite{hartvigsen2022toxigen} and ToxiCraft~\cite{hui2024toxicraft} generate adversarial datasets to stress-test models. Studies~\cite{luong2024realistic,deshpande2023toxicity,chetnani2023evaluating,oh2024uniguard} examine the relationship between model size and toxicity generation, finding that smaller models often exhibit lower toxicity rates.

\paratitle{Human-AI Collaboration.} Research demonstrates that human intervention significantly improves toxicity detection quality~\cite{sen2023people}, particularly through counterfactual data augmentation. Additional work explores covert toxicity detection~\cite{lee2024improving}, data contamination~\cite{balloccu2024leak}, and geometric interpretability~\cite{balestrierocharacterizing} to enhance model safety.

\subsection{Distribution Stabilization}\label{sec:distribution_stabilization}
Stabilizing data distribution ensures that models generalize well across different tasks and domains. Several methods tackle issues like class imbalance, noise reduction, and domain adaptation:

\paratitle{Imbalance Mitigation.} Approaches like Synthetic Oversampling~\cite{nakada2024synthetic} and Diversify and Conquer~\cite{yu2024diversify} effectively address class imbalance through adaptive synthetic sample generation. Studies show significant improvements, with~\cite{cai2023resolving} demonstrating a 38\% fairness boost in cross-disciplinary applications.

\paratitle{Noise Reduction.} Multi-News+~\cite{choi2024multi} signficantly reduces annotation errors through automated label correction, while~\cite{chen2024automated} employs self-supervised filtering for robust fine-tuning data curation. RobustFT~\cite{luo2024robustft} introduces a comprehensive framework for handling noisy response data through multi-expert collaborative noise detection and context-enhanced relabeling strategies, coupled with entropy-based data selection for high-quality sample retention.

\paratitle{Domain Adaptation.} ChatTS~\cite{xie2024chatts} uses Fourier transforms for time-series alignment, while~\cite{becker2024conditional} applies domain-specific label smoothing for clinical text. Advanced approaches like Dynosaur~\cite{yin2023dynosaur} and Optima~\cite{chen2024optima} leverage curriculum learning and multi-source coordination. RADA~\cite{seo2024retrieval} addresses low-resource domain tasks by retrieving relevant instances from other datasets and generating contextually enhanced samples through LLM prompting.

\subsection{Discussion}
Semantic rewriting, toxicity control, and distribution stabilization represent key strategies for improving data quality in LLM post-training. These techniques ensure the generation of diverse, high-quality data, mitigate harmful content, and stabilize data distributions to enhance model robustness. Future work should integrate these approaches into unified frameworks to maximize data diversity and model performance while reducing costs.

% \subsection{Discussion}
% Semantic rewriting, toxicity control, and distribution stabilization represent key strategies for improving data quality in LLM post-training. While each method addresses specific challenges, future research should focus on developing integrated solutions that combine these approaches efficiently, balancing quality improvements with compute costs.

\section{Synthetic Data Generation}\label{sec:synthetic_data_generation}
\begin{figure}[t]
    \centering
    \includegraphics[width=\linewidth]{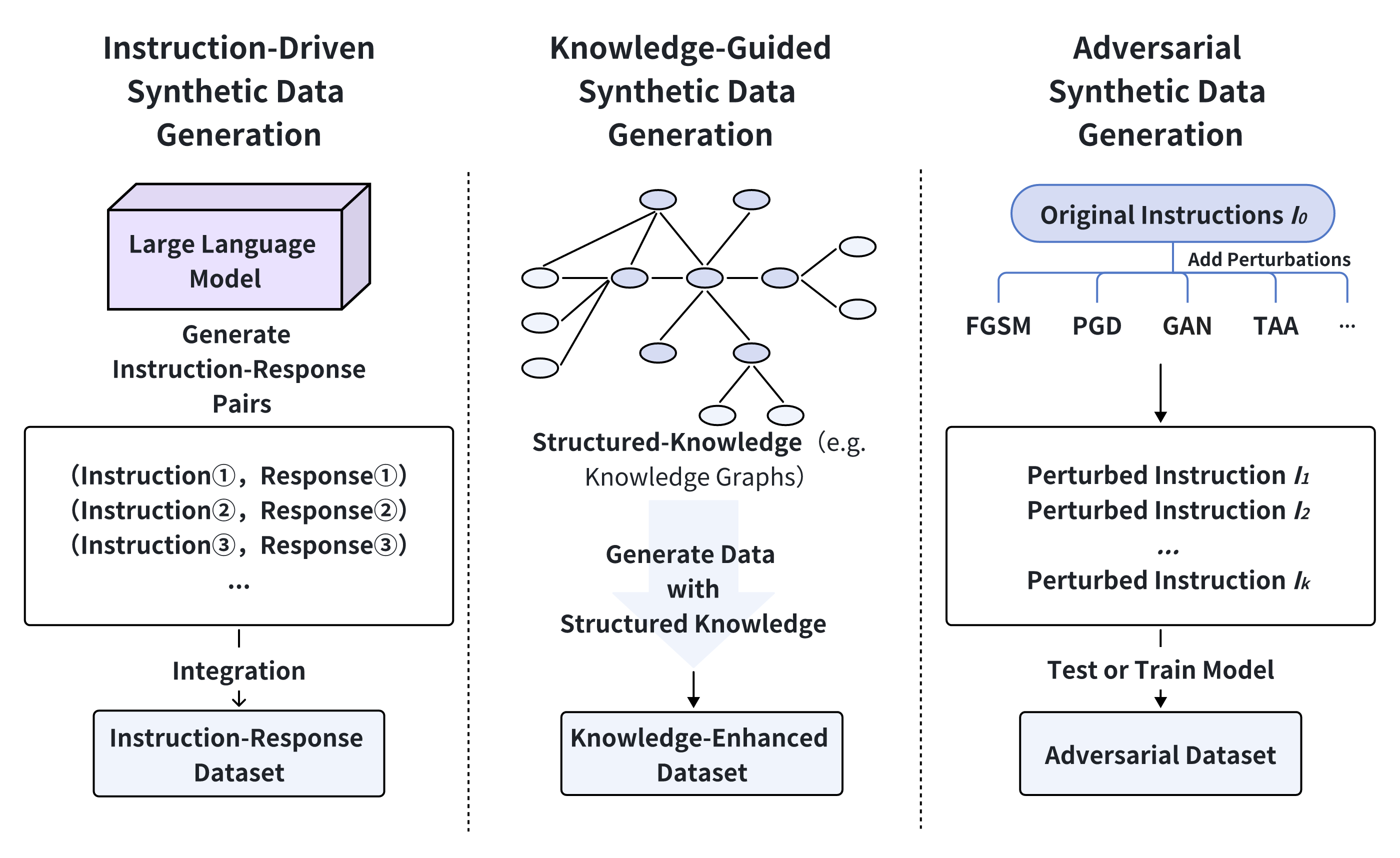}
    \vspace{-5mm}
    \caption{Three main approaches for data generation in LLM post-training: instruction-driven generation for creating instruction-response pairs, knowledge-guided generation using structured knowledge, and adversarial generation for testing model robustness.}
    \vspace{-5mm}
    \label{fig:data-generation}
\end{figure}

Generating synthetic training data is a powerful strategy to overcome data scarcity and enhance the robustness of LLM post-training. As illustrated in Figure~\ref{fig:data-generation}, synthetic data generation methods can be categorized into three main approaches: \textit{Instruction-Driven}, \textit{Knowledge-Guided}, and \textit{Adversarial Generation}, each serving distinct purposes in enhancing model capabilities.

\subsection{Instruction-Driven Synthetic Data Generation}\label{sec:instruction_driven_synthetic_data_generation}
Instruction-driven methods harness LLMs' ability to produce new examples directly from task prompts.
Recent works demonstrate diverse applications: SynPO~\cite{dong2024self} generates preference pairs for alignment (12\% ROUGE-L improvement), Magpie~\cite{xu2024magpie} enables template-free instruction generation (98\% AlpacaEval accuracy), and Advancing Theorem Proving~\cite{xin24advancing} synthesizes Lean4 proof steps, boosting GPT-4's proving capabilities by 34\%.

\subsection{Knowledge-Guided Synthetic Data Generation}\label{sec:knowledge_guided_synthetic_data_generation}
Knowledge-guided approaches integrate external knowledge to steer data generation.

\paratitle{Theoretical Frameworks.} Towards a Theoretical Understanding~\cite{gan2024towards} rigorously establishes a reverse-bottleneck theory linking data diversity to enhanced model generalization.

\paratitle{Structured Data Synthesis.} HARMONIC~\cite{wang2024harmonic} combines privacy-preserving tabular data generation on medical records.~\cite{xu2024llms} improves relational consistency through schema-aware fine-tuning.

\paratitle{Cost-Effective Strategies.}~\cite{chan2024balancing} demonstrates hybrid generation methods reduce API costs by 70\% while maintaining data utility. Source2Synth~\cite{lupidi2024source2synth} improves factual accuracy through knowledge-graph alignment.

\subsection{Adversarial Generation}\label{sec:adversarial_generation}
Adversarial generation methods systematically probe model vulnerabilities to enhance robustness. Recent works demonstrate diverse approaches: Illuminating Blind Spots~\cite{lippmann2024illuminating} uses agent-based simulations to generate edge cases, reducing errors by 19\% on dialect variation; Unveiling Synthetic Data Flaws~\cite{chen2024unveiling} introduces contrastive unlearning to address data imperfections, yielding 32\% quality improvements on GLUE; and ToxiCraft~\cite{hui2024toxicraft} generates subtle harmful content, revealing significant gaps in commercial safety filters.

\subsection{Discussion}
Each approach offers distinct trade-offs: instruction-driven methods enable rapid scaling but risk semantic drift; knowledge-guided approaches maintain fidelity through structured constraints; and adversarial generation strengthens robustness by exposing vulnerabilities. Future work should combine these strengths—for instance, merging privacy-preserving generation with adversarial testing. Key challenges persist in optimizing generation costs~\cite{chan2024balancing} and developing theoretical foundations~\cite{gan2024towards}.
\section{Data Distillation and Compression}\label{sec:data_distillation_compression}

\begin{figure}[t]
    \centering
    \includegraphics[width=\linewidth]{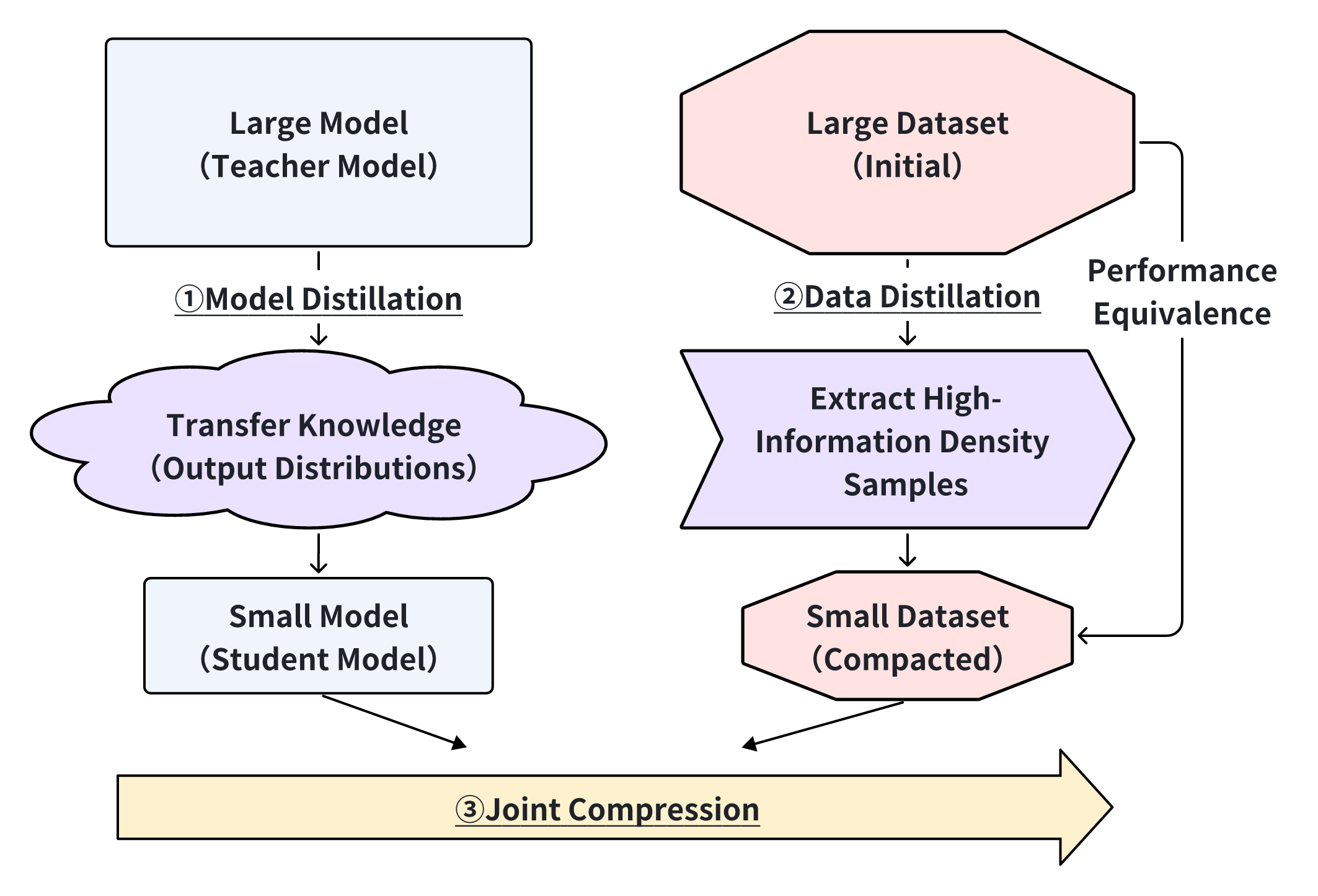}
    \vspace{-5mm}
    \caption{Data distillation and compression in LLM post-training: model distillation for knowledge transfer, data distillation for sample extraction, and joint compression for unified optimization.}
    \vspace{-5mm}
    \label{fig:data-distillation}
\end{figure}

Data distillation and compression techniques enhance LLM post-training efficiency by reducing data complexity while preserving performance. As shown in Figure~\ref{fig:data-distillation}, this involves three complementary approaches: model distillation for knowledge transfer, data distillation for dataset compression, and joint compression for unified optimization.

\subsection{Model Distillation}\label{sec:model_distillation}

Model distillation transfers knowledge from large to smaller models while maintaining performance. Recent advances include Impossible Distillation~\cite{jung2023impossible}, which creates high-quality models from low-quality teachers, and Performance-Guided Distillation~\cite{di2024performance}, achieving 98\% accuracy with 40\% reduced costs. Cross-Tokenizer Distillation~\cite{boizard2024towards} enables knowledge transfer between different architectures through universal logit distillation. For edge deployment, XAI-Driven Distillation~\cite{cantini2024xai} produces interpretable medical models, while BitDistiller~\cite{du2024bitdistiller} enables sub-4-bit precision with minimal accuracy loss. Multistage Collaborative Distillation~\cite{zhao2024multistage} improves performance through multi-teacher coordination in low-resource settings.

\subsection{Data Distillation}\label{sec:data_distillation}

Data distillation focuses on selecting high-information-density samples to create compact yet representative datasets. Knowledge Distillation in Automated Annotation~\cite{pangakis2024knowledge} shows LLM-generated labels can effectively train classifiers comparable to human annotations. LLMLingua-2~\cite{pan2024llmlingua} approaches prompt compression through token-level distillation. Domain-specific applications include Self-Data Distillation~\cite{thangarasa2024self} for model refinement, Multi-Teacher Distillation~\cite{zhang2024llm} for healthcare data integration, and techniques for reducing hallucination~\cite{mcdonald2024reducing}.

\subsection{Joint Compression}\label{sec:joint_compression}

Joint compression combines model compression with data selection to optimize overall efficiency. Compact Language Models via Pruning and Distillation~\cite{muralidharan2024compact} co-optimizes structural pruning and label smoothing, compressing LLaMA-7B to 2.8B parameters with minimal performance loss. Efficient Edge Distillation~\cite{cantini2024xai} enables adaptive width scaling for edge devices through supernet training. In recommendation systems, Prompt Distillation~\cite{li2023prompt} aligns ID-based and text-based representations, aiming to reduce inference time.

For multimodal applications, recent work demonstrates joint compression of graph and text encoders~\cite{pan2024distilling} and curriculum-aligned prompt distillation for educational LLMs~\cite{qu2024coursegpt}, achieving significant parameter reduction while maintaining performance.

\subsection{Discussion}
These three approaches offer complementary benefits for enhancing LLM efficiency: model distillation optimizes architecture, data distillation curates high-impact samples, and joint compression unifies model-data optimization. Future research should focus on integrating these methods, particularly for edge AI and low-resource applications.

\begin{figure}[t]
    \centering
    \includegraphics[width=0.8\linewidth]{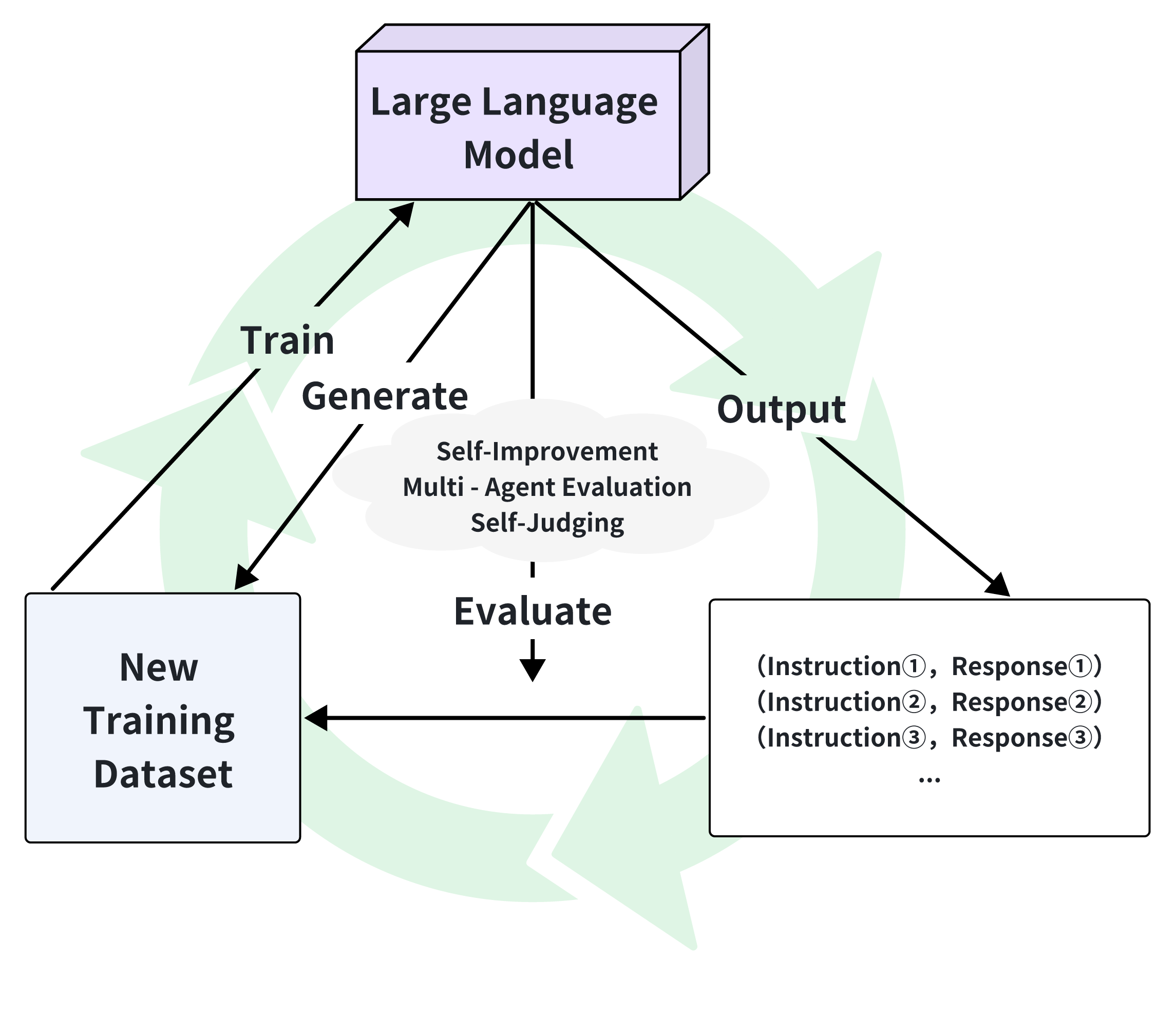}
    \vspace{-5mm}
    \caption{Self-Evolving Data Ecosystem: autonomous data generation, real-time feedback, and continuous learning.}
    \vspace{-5mm}
    \label{fig:selfevol}
\end{figure}

\section{Self-Evolving Data Ecosystem}\label{sec:self_evolving}
The Self-Evolving Data Ecosystem strategically optimizes LLM post-training through autonomous data generation, real-time feedback, and continuous learning. As shown in Figure~\ref{fig:selfevol}, this ecosystem forms a closed loop of generation, evaluation, and adaptive training. We discuss three key components: Self-Iterative Optimization, Dynamic Evaluation Feedback, and LLM-as-a-Judge.

\subsection{Self-Iterative Optimization}\label{sec:self_iterative_optimization}
Self-iterative optimization enables LLMs to use their own outputs to generate new training data, refining their capabilities autonomously. Several approaches illustrate this concept:

\paratitle{Self-Improvement Methods.} Recent works like Self-Rewarding~\cite{yuan2024self}, Self-Refine~\cite{madaan2024self}, and Self-Boosting~\cite{dong2024self} enable models to autonomously improve through iterative self-optimization. Self-Play Fine-Tuning~\cite{chen2024self} extends this by leveraging competitive self-interaction, outperforming traditional methods like DPO~\cite{rafailov2024direct}.

\paratitle{Semi-Supervised Self-Evolution.} In practical deployment scenarios, models often encounter limited labeled seed data alongside abundant unlabeled domain-specific data, creating a critical challenge for effective post-training adaptation. SemiEvol~\cite{luo2025semi} addresses this challenge through a propagate-and-select framework that transfers knowledge from seed data to unlabeled samples via bi-level propagation and collaborative selection mechanisms.

\paratitle{Knowledge Retention.} In the context of retaining knowledge while integrating new data, MemoryLLM~\cite{wang2024memoryllm} enables continuous model updates while preserving existing knowledge. Automated Proof Generation~\cite{chen2024automated} and Arxiv Copilot~\cite{lin2024arxiv} demonstrate this capability in code verification and academic research tasks.

\subsection{Dynamic Evaluation Feedback}\label{sec:dynamic_evaluation_feedback}
Dynamic evaluation feedback systems allow models to make real-time adjustments based on their performance, optimizing their outputs on the fly. Key contributions include:

\paratitle{Multi-Agent Evaluation.} The Benchmark Self-Evolving Framework~\cite{wang2024benchmark} and LLM-Evolve~\cite{you2024llm} employ multi-agent systems to evaluate and adjust LLM performance dynamically. These frameworks enable the models to self-adjust in real-time across various benchmarks, ensuring continuous evolution.

\paratitle{Iterative Refinement} Self-Refine~\cite{madaan2024self} and Self-Log~\cite{pei2024self} employ feedback loops for iterative refinement and log parsing, optimizing the model's output without requiring external retraining.I-SHEEP~\cite{liang2024sheep} offers a resource-efficient paradigm that enhances performance through self-alignment, while Interactive Evolution: A Neural-Symbolic Self-Training Framework~\cite{xu2024interactive} enables LLMs to autonomously train in neural-symbolic environments.

\paratitle{Improved Decision Making.} For improving model alignment, Meta-Rewarding~\cite{wu2024meta} and Self-Evolved Reward Learning~\cite{huang2024self} leverage iterative feedback from their outputs to improve judgment skills, ensuring more accurate decision-making in complex tasks.

\subsection{LLM-as-a-Judge}\label{sec:llm_as_a_judge}
LLM-as-a-Judge systems represent a paradigm shift from external evaluation to self-assessment, where models evaluate their own or other models' outputs. These systems operate through three fundamental mechanisms, each addressing different evaluation challenges:

\paratitle{Self-Improvement through Judgment.} These methods focus on improving a model's ability to assess quality. Self-Taught Evaluators~\cite{wang2024self} and Meta-Rewarding~\cite{wu2024meta} take distinct approaches: the former generates synthetic comparisons to train judgment without human data, while the latter introduces meta-judgment by having models critique their own evaluations. JudgeLM~\cite{zhu2023judgelm} takes a different path by fine-tuning models on human preferences to create specialized evaluation models.

\paratitle{Debiasing Evaluation Systems.} These methods address fairness concerns in automated evaluation. CalibraEval~\cite{li2024calibraeval} recalibrates prediction distributions to mitigate position bias, while Crowd Score~\cite{goes2022crowd} employs multiple AI \textit{personalities} within a single model to simulate diverse human judgments, reducing individual bias through aggregation.

\paratitle{Adversarial Robustness Testing.} These approaches stress-test models through challenging scenarios. TOXIGEN~\cite{hartvigsen2022toxigen} and ToxiCraft~\cite{hui2024toxicraft} create progressively more subtle toxic content to expose blind spots, while R-Judge~\cite{yuan2024r} specifically targets situational safety risks in interactive environments rather than just content harmfulness.

\subsection{Discussion}
The combination of Self-Iterative Optimization, Dynamic Evaluation Feedback, and LLM-as-a-Judge creates a robust framework for autonomous LLM improvement. While these approaches show promise in reducing manual intervention, future work should focus on unifying them into scalable frameworks that generalize across diverse tasks.
\section{Challenges and Future Directions}\label{sec:future_work}

\paratitle{Domain-Driven Data Synthesis and Refinement.}
While general-purpose models like GPT are commonly used for data generation~\cite{di2024performance}, domain-specific models can better capture professional knowledge~\cite{lightman2023let}. Future work should explore domain-specific pre-trained models for generating specialized data~\cite{luo2023wizardcoder,cheng2024adapting}, along with refinement techniques to optimize data quality while reducing annotation costs.

\paratitle{Scalability of Large-Scale Data Synthesis.}
As LLM pre-training demands increasingly larger and higher-quality datasets, efficient large-scale data generation becomes crucial. Current data synthesis and augmentation methods face scalability bottlenecks. Future work should focus on developing parallel, cost-effective, and efficient data generation frameworks that meet the demands of large-scale pre-training while maintaining a balance between data diversity and relevance~\cite{karunya2023ai}.

\paratitle{Reliable Quality Assessment Metrics.}
Current evaluation frameworks lack standardized metrics for assessing synthetic data quality~\cite{zendel2024enhancing}. Future research should develop metrics that evaluate semantic fluency, information accuracy, and potential biases~\cite{chundawat2022universal,gerstgrasser2024model} to ensure robust selection.

\section{Conclusion}

In this paper, we presented a systematic review of LLM post-training research from a data efficiency perspective. We established the first taxonomic framework for data-efficient post-training, encompassing five core methodologies. Through detailed analysis of representative approaches within each category, we revealed that breaking through data efficiency bottlenecks requires establishing value extraction mechanisms across the entire data lifecycle. We aimed to highlight the current state and provide valuable insights for future work in this promising field of data-efficient LLM post-training.

\subsection*{Limitations}

While our work presents the first comprehensive framework for analyzing data-efficient LLM post-training approaches, several limitations and opportunities for future research remain. First, given the explosive growth of this field, some emerging techniques may not be fully captured in our current taxonomic system, necessitating continuous updates to maintain comprehensiveness. Second, while data efficiency is crucial, the proposed methods may face additional challenges regarding trustworthiness and scalability that warrant further investigation. Furthermore, the synergistic effects and interaction mechanisms between different data efficiency enhancement techniques remain underexplored, calling for the development of cross-method optimization theories. We anticipate these open challenges will inspire deeper theoretical innovations and practical breakthroughs.

% \clearpage

\section*{Acknowledgments}

This paper is partially supported by grants from the National Key Research and Development Program of China with Grant No. 2023YFC3341203 and the National Natural Science Foundation of China (NSFC Grant Number 62276002). The authors are grateful to the anonymous reviewers for their efforts and insightful suggestions to improve this article.

% \clearpage

% \clearpage

\bibliography{custom}

% \clearpage

% \newpage

\appendix
\label{appendix}

\section{Statistics}

\begin{figure}[h]
    \centering
    \includegraphics[width=\linewidth]{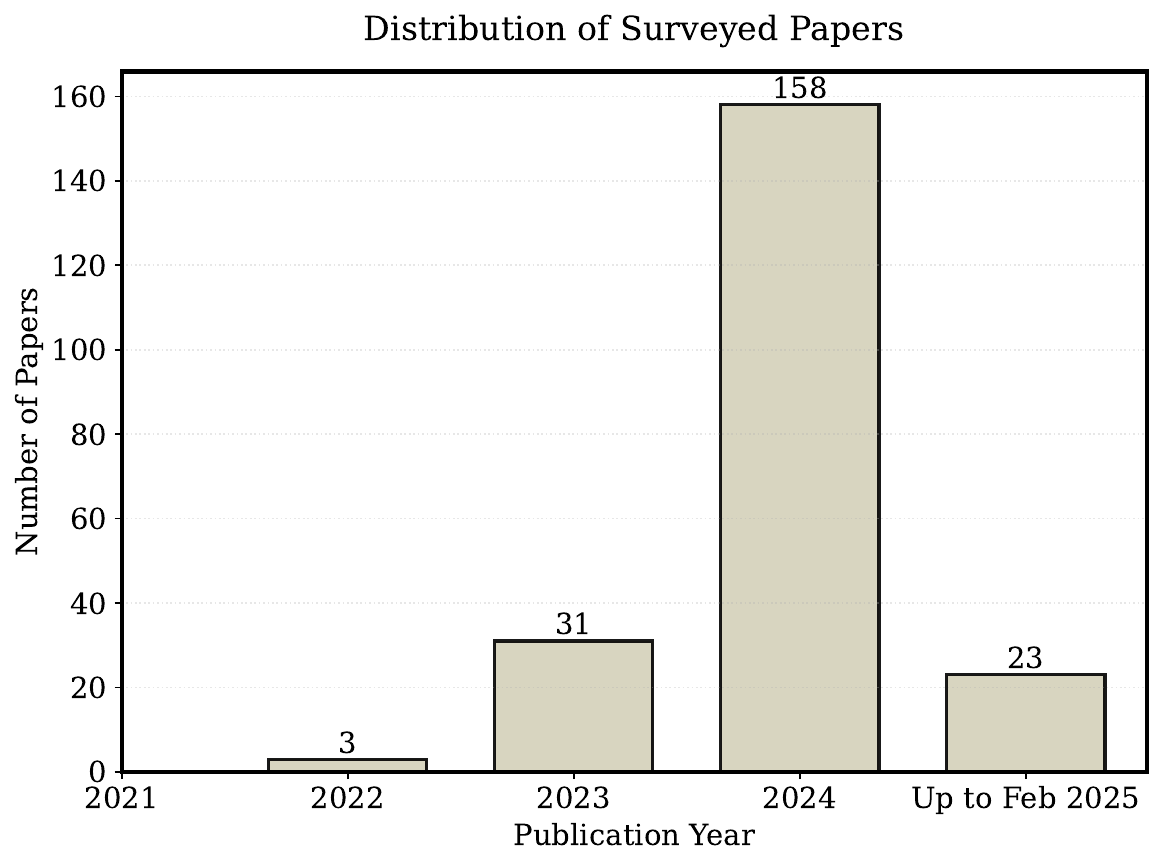}
    \caption{Distribution on publication year of surveyed papers.}
    \label{fig:year_distribution}
\end{figure}

To demonstrate the research momentum in data-efficient LLM post-training, we conducted a statistical analysis of the surveyed papers. As shown in Figure~\ref{fig:year_distribution}, there has been a remarkable growth trajectory in this field: from merely $3$ publications in $2022$ to $31$ papers in $2023$, followed by a substantial surge to $158$ papers in $2024$, with $23$ additional publications already recorded by February $2025$. This trend clearly indicates the academic community's growing interest in this research direction, with the momentum continuing to accelerate. The rapid growth also underscores the critical importance of data-efficient post-training approaches in the LLM domain.

\begin{figure}[h]
    \centering
    \includegraphics[width=\linewidth]{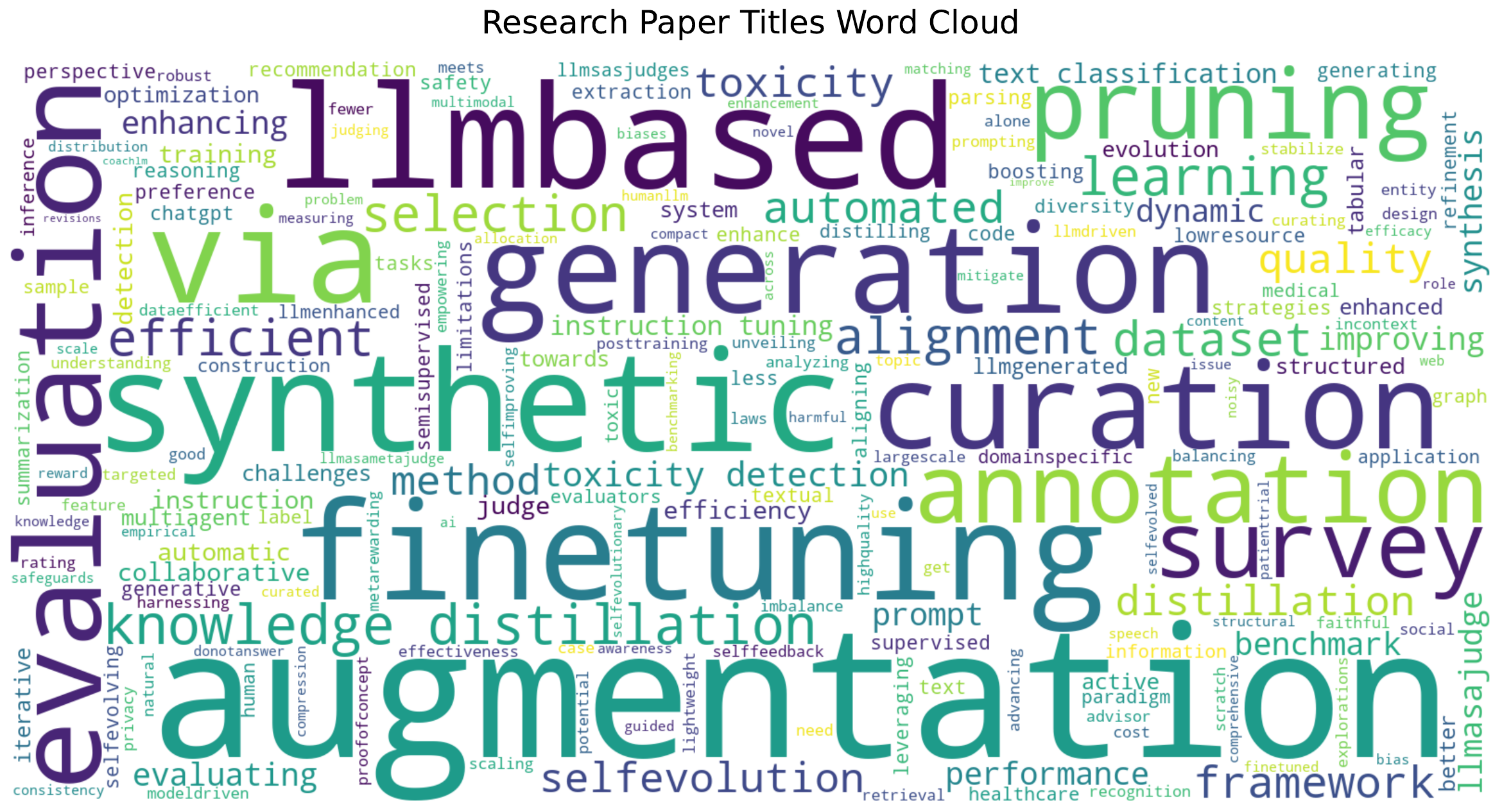}
    \caption{Word cloud of research paper titles.}
    \label{fig:wordcloud}
\end{figure}

Furthermore, we performed a word frequency analysis on the titles of all surveyed papers and generated a word cloud visualization (Figure~\ref{fig:wordcloud}). The word cloud reveals key methodological focal points in current research, with \textit{augmentation}, \textit{synthetic}, \textit{generation}, and \textit{alignment} emerging as prominent themes. The significant presence of terms like \textit{finetuning}, \textit{distillation}, and \textit{efficient} underscores the field's emphasis on optimizing model training processes. These visualizations demonstrate the centrality of data-centric approaches and synthetic data methodologies in advancing LLM post-training efficiency.

The analysis demonstrates that data-efficient approaches to LLM post-training represent not only an emerging trend but also a fundamental research direction with significant implications for the advancement of language models.

\section{Takeaway Insights}
\label{sec:insights}

\subsection{Key Findings} \label{A.1}
Recent advancements in data-efficient LLM post-training reveal fundamental principles governing data-model interactions:
\begin{itemize}
    \item[(1)] The \textit{\textbf{data flywheel}} paradigm integrates selection, augmentation, and evolution into a closed-loop lifecycle. This self-reinforcing mechanism enables continuous quality improvement through iterative refinement, transcending traditional linear data consumption
    
    \item[(2)] \textbf{Value-centric data curation} outperforms scale-driven approaches in low-resource scenarios. Techniques like adaptive importance weighting and uncertainty-aware sampling maximize information density per training instance
    
    \item[(3)] \textbf{Model-data co-optimization} enables joint improvements in efficiency and performance through innovations like dynamic token pruning and parameter-efficient adaptation
\end{itemize}

\subsection{Paradigm Shifts} \label{A.2}
The field is witnessing fundamental changes in data utilization:
\begin{itemize}
    \item[(1)] Evolution from static datasets to \textbf{dynamic value-flow ecosystems} where data continuously evolves through model feedback. This necessitates new frameworks for monitoring data quality and lineage across iterations
    
    \item[(2)] Emergence of \textbf{human-AI collaborative frameworks} combining automated generation with expert oversight. These hybrid pipelines leverage LLMs for initial labeling while preserving human judgment for critical cases
    
    \item[(3)] Development of \textbf{cross-modal distillation} techniques that maintain semantic fidelity while reducing architectural constraints through learned alignment spaces
\end{itemize}

\subsection{Critical Limitations} \label{A.3}
Current approaches face several key challenges:
\begin{itemize}
    \item[(1)] \textbf{Limited domain expertise} in data synthesis and refinement, where general-purpose models may fail to capture specialized knowledge and nuances required for professional domains
    
    \item[(2)] \textbf{Scalability bottlenecks} in large-scale data generation, particularly in balancing computational costs with the need for diverse, high-quality datasets for pre-training
    
    \item[(3)] Absence of \textbf{standardized metrics} for assessing synthetic data quality, especially in evaluating semantic fluency, information accuracy, and potential biases
\end{itemize}

\subsection{Future Directions} \label{A.4}
Addressing these limitations requires advances in:
\begin{itemize}
    \item[(1)] \textbf{Domain-specific} pre-trained models and refinement techniques that can better capture professional knowledge while optimizing data quality and reducing annotation costs
    
    \item[(2)] \textbf{Parallel and cost-effective frameworks} for large-scale data generation that maintain an optimal balance between data diversity and relevance
    
    \item[(3)] \textbf{Robust evaluation metrics} and frameworks that can reliably assess synthetic data quality across different domains and use cases
\end{itemize}

\section{Acknowledgment of AI Assistance in Writing and Revision}
We acknowledge the use of LLMs for grammar checking and language enhancement. This usage complies with the ACL Policy on AI Writing Assistance. All content and technical contributions remain original to the authors.

\section{Literature Review Summary}

To provide a comprehensive overview of the surveyed literature, we present a detailed summary table of all referenced papers. The table includes seven key fields for each paper: \textbf{Title} (the paper's full title), \textbf{Citation} (reference key), \textbf{TLDR} (a brief summary of the paper's main contributions), \textbf{Category} (the paper's primary research direction within data-efficient LLM post-training), \textbf{Year} (publication year), \textbf{Venue} (publication venue), and \textbf{Link} (direct link to the paper). This structured compilation offers readers quick access to the original papers, enables easy tracking of research evolution across different categories, and facilitates future research by providing a comprehensive reference database of the field's development.

\end{document}